\def\year{2021}\relax
\documentclass[letterpaper]{article} 
\usepackage{aaai21}  
\usepackage{times}  
\usepackage{helvet} 
\usepackage{courier}  
\usepackage[hyphens]{url}  
\usepackage{graphicx} 
\usepackage{natbib}  
\usepackage{caption} 
\newcommand{\ie}{\textit{i}.\textit{e}.}

\usepackage{booktabs}       
\usepackage{amsfonts}       
\usepackage{amsmath}
\usepackage{ dsfont }

\usepackage{paralist}
\usepackage{multirow}
\usepackage{graphicx}
\usepackage{subcaption}
\usepackage[ruled]{algorithm2e}
\SetKwComment{Comment}{$\triangleright$\ }{}
\usepackage{tikz}
\usetikzlibrary{positioning}
\usepackage{soul}

\title{Universal Adversarial Perturbations Through the Lens of Deep Steganography: Towards A Fourier Perspective}
\author{
    Chaoning Zhang$^*$,
    Philipp Benz$^*$,
    Adil Karjauv,
    In So Kweon\\
}
\affiliations{
    Korea Advanced Institute of Science and Technology (KAIST)\\
    chaoningzhang1990@gmail.com, pbenz@kaist.ac.kr, mikolez@gmail.com, iskweon77@kaist.ac.kr\\
    $^*$ Equal contribution

}

\begin{document}

\maketitle

\begin{abstract}
The booming interest in adversarial attacks stems from a misalignment between human vision and a deep neural network (DNN), \ie~a human imperceptible perturbation fools the DNN. Moreover, a single perturbation, often called universal adversarial perturbation (UAP), can be generated to fool the DNN for most images. A similar misalignment phenomenon has recently also been observed in the deep steganography task, where a decoder network can retrieve a secret image back from a slightly perturbed cover image. We attempt explaining the success of both in a unified manner from the Fourier perspective. We perform task-specific and joint analysis and reveal that (a) frequency is a key factor that influences their performance based on the proposed entropy metric for quantifying the frequency distribution; (b) their success can be attributed to a DNN being highly sensitive to high-frequency content. We also perform feature layer analysis for providing deep insight on model generalization and robustness. Additionally, we propose two new variants of universal perturbations: (1) Universal Secret Adversarial Perturbation (USAP) that simultaneously achieves attack and hiding; (2) high-pass UAP (HP-UAP) that is less visible to the human eye. \footnote{Supplementary can be found at \url{chaoningzhang.github.io/publication/aaai2021/supplementary.pdf}}
\end{abstract}

\section{Introduction}
\label{intro}

\begin{figure*}[!htbp]
\centering
    \includegraphics[width=\linewidth]{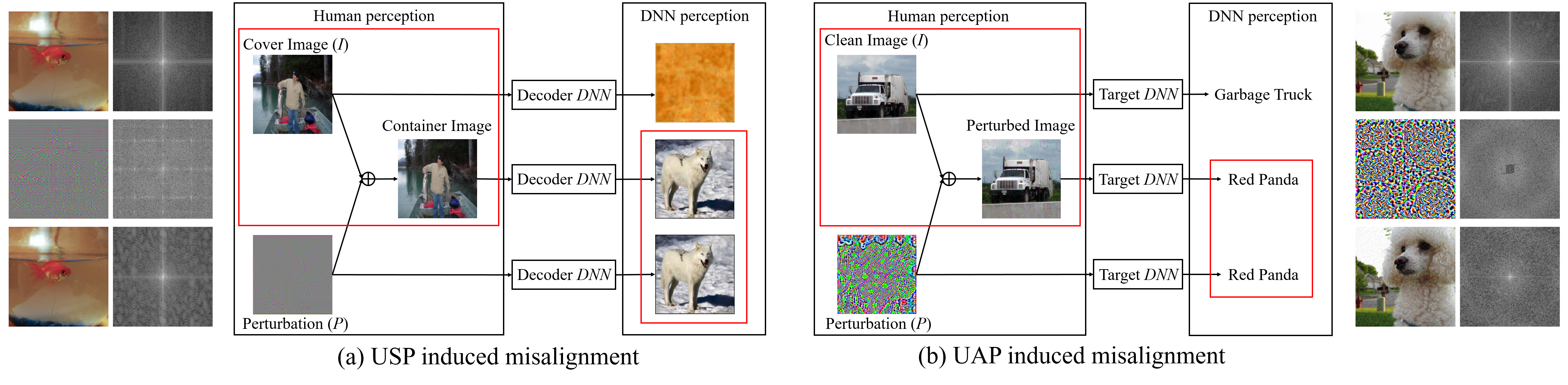}
    \caption{Misalignment under the universal framework. (a) USP induced misalignment; (b) UAP induced misalignment. In both (a) and (b): given $||P||\ll||I||$, $H(I+P)\approx H(I)$ while $M(I+P)\approx M(P)$. To both sides example images and their Fourier images for the respective task are shown. From top to bottom the images represent: clean image ($I$), amplified perturbation ($P$), and perturbed image ($I+P$). The corresponding Fourier images show that $P$ has HF property contrary to that of $I$.}
    \label{fig:main_framework}
\end{figure*}

Deep learning has achieved large success in a wide range of vision applications, such as recognition~\cite{zhang2019revisiting,zhang2020resnet}, segmentation~\cite{vania2019automatic,kim2020video,pan2020unsupervised} as well as scene understanding~\cite{lee2019learning,lee2019visuomotor,zhang2020deepptz,argaw2021motion,argaw2021optical}. Nonetheless, the vulnerability of deep neural networks (DNNs) to adversarial examples~\cite{szegedy2013intriguing} has attracted significant attention in recent years. In machine learning, there is a surging interest in understanding the reason for the success of the adversarial attack (AA)~\cite{szegedy2013intriguing,zhang2020understanding}. The root reason for this booming interest lies in the misalignment between human vision and DNN perception (see Figure~\ref{fig:main_framework}).
A similar misalignment phenomenon has also been observed in deep steganography (DS)~\cite{baluja2017hiding,zhang2020udh}, where a decoder network retrieves a secret image from a slightly perturbed cover image, often referred to as container image. In this work, for consistency, a small change to an image is termed perturbation ($P$) for both DS and AA. In both tasks, the original image $I$ and perturbed image $I+P$ are nearly indistinguishable for the human vision system, given $||P||\ll||I||$ (see Figure~\ref{fig:main_framework}). However, for a DNN, $M(I+P)$ is more similar to $M(P)$ than $M(I)$ where $M$ indicates the model of interest as a function. For AA and DS, the DNN of interest is the target DNN and decoder network, respectively. For an instance-dependent perturbation (IDP) case, taking AA for example, this misalignment is relatively less surprising. We focus on the misalignment in ``universal" scenario, with conflicting features in $I$ and $P$, while $I$ is dominated by $P$ when they are summed, \ie\ $I+P$, as the $M$ input.

For both AA and DS, the misalignment constitutes the most fundamental concern, thus we deem it insightful to explore them together. We first attempt explaining its misalignment based on our adopted universal secret perturbation (USP) generation framework introduced in~\cite{zhang2020udh}, where a secret image is hidden in a cover-agnostic manner. The success of DS has been attributed to the discrepancy between $C$ and the encoded secret image~\cite{zhang2020udh}. Inspired by the success of explaining the USP induced misalignment from the Fourier perspective, we explore the UAP induced misalignment in a similar manner.

Our analysis shows that the influence of each input on the combined DNN output is determined by both frequency and magnitude, but mainly by the frequency. To quantitatively analyze the influence of image frequency on the performance of the two tasks, we propose a new metric for quantifying the frequency that involves no hyperparameter choices. Overall, our task-specific and cross-task analysis suggest that image frequency is a key factor for both tasks. 

Contrary to prior findings regarding IDP in~\cite{yin2019fourier}, we find that UAPs, which attack most images are a strictly high-frequency (HF) phenomenon. Moreover, we perform a feature layer analysis to provide insight on model generalization and robustness. With the frequency understanding, we propose two novel universal attack methods. 

\section{Related work}

\textbf{Fourier perspective on DNN.} 
The behavior of DNNs has been explored from the Fourier perspective in multiple prior arts. Some works~\cite{jo2017measuring,wang2020high} analyze why the DNN has good generalization while being vulnerable to adversarial examples. Their results suggest that surface-statistical regularities, exhibiting HF property, are useful for classification. Similar findings have also been shown in~\cite{ilyas2019adversarial} that human unrecognizable non-robust-features with HF property are sufficient for the model to exhibit high generalization capability. On the other hand, DNNs trained only on low-pass filtered images appearing to be simple globs of color are also found to be sufficient for generalizing with high accuracy~\cite{yin2019fourier}. Overall, there is solid evidence that both HF features and LF features can be useful for classification. It is interesting to explore whether a DNN is more biased towards HF or LF features. One work~\cite{geirhos2018imagenet} shows that DNNs are more biased towards texture than shape through a texture-shape cue conflict. Given that texture mainly has HF content and the shape can be seen to have LF content (most flat regions except the object boundary), it can be naturally conjectured that DNNs are more biased towards HF content. We verify this by presenting extensive analysis. We acknowledge that this does not constitute a major discovery, instead, we highlight that \textit{we apply it to explain the model robustness to UAPs in the context of independent yet conflicting features in the $I+P$}. 

Regarding the Fourier perspective to model robustness, adversarial perturbations are widely known to have the HF property, motivated by which several defense methods~\cite{aydemir2018effects,das2018shield,liu2019feature} have been explored. However, \citeauthor{yin2019fourier} concluded that ``\textit{Adversarial examples are not strictly a high frequency phenomenon}", which echoed with explorations of LF perturbations~\cite{guo2020low,sharma2019effectiveness} as well as the finding in~\cite{carlini2017adversarial} regarding false claims of detection methods that use PCA~\cite{gong2017adversarial,grosse2017statistical,metzen2017detecting}. Our claim that \textit{UAPs attacking most images is a strictly HF phenomenon} does not conflict with the claim in~\cite{yin2019fourier} because they implicitly mainly discuss IDPs, not UAPs.

\textbf{On universal adversarial attack.} The reason for the existence of IDP has been analyzed from various perspectives~\cite{qiu2019review}, such as local linearity~\cite{goodfellow2014explaining, tabacof2016exploring}, input high-dimension~\cite{shafahi2018adversarial,fawzi2018adversarial,mahloujifar2019curse,gilmer2018adversarial}, limited sample~\cite{schmidt2018adversarially,tanay2016boundary}, boundary tilting~\cite{tanay2016boundary}, test error in noise~\cite{fawzi2016robustness,ford2019adversarial,cohen2019certified}, non-robust features~\cite{bubeck2019adversarial,nakkiran2019a,ilyas2019adversarial}, batch normalization~\cite{benz2020revisiting,benz2020batch} etc. These explanations for IDPs do not come to a consensus that can be directly used to explain the existence of UAPs. The image-agnostic nature of UAPs requires a specific explanation. Relevant analysis has been performed in~\cite{moosavi2017universal, moosavi2017analysis, jetley2018friends, moosavi2019robustness}. Their analysis focused on why a single UAP can fool most samples across the decision boundary and they attributed the existence of UAPs to the large curvature of the decision boundary.~\cite{zhang2020understanding} shows that UAPs have independent semantic features that dominate the image features. In this work, we analyze the role of frequency in images being dominated by the UAP. Recently, class-wise UAPs~\cite{zhang2019cd-uap} and double targeted UAPs~\cite{benz2020double} have also been investigated for making the universal attack more stealthy. 
  
\textbf{When adversarial examples meet deep steganography.} Deep hiding has recently become an active research field. Hiding binary messages has been explored in~\cite{hayes2017generating,zhu2018hidden,wengrowski2019light} and hiding image (or videos) has been explored in~\cite{baluja2017hiding,weng2018convolutional,mishravstegnet}. Interpretability of DNNs has become one important research direction, thus it is also crucial to understand how the DNN works in DS.~\cite{baluja2017hiding,baluja2019hiding} disproves the possibility of the secret image being hidden in the least significant bit (LSB). Recent work~\cite{zhang2020udh} shows that the success of DS can be attributed to the frequency discrepancy between cover image and encoded secret image. Joint investigation of AA and DS has also been investigated by proposing a unified notion of black-box attacks against both tasks~\cite{quiring2018forgotten}, applying the lesson in multimedia forensics to detect adversarial examples~\cite{schottle2018detecting}. Our work differentiates by focusing on the ``universal" property with a Fourier perspective.

\section{Motivation and background prior}

\textbf{Why studying AA and DS together with universal perturbation?} 
Technically, UAPs are crafted to attack a target DNN while DS learns a pair of DNNs for encoding/decoding. 
Both tasks share a misalignment phenomenon between the human observer and the involved DNN. Specifically, in both cases, a human observer finds that the perturbed image looks natural, but the DNN gets fooled (for AA) or reveals a hidden image (for DS). Motivated by the observation of shared misalignment phenomenon, we deem it meaningful to study the two tasks in parallel to provide a unified perspective on this phenomenon. Moreover, studying them together allows us to perform cross-task analysis which can further strengthen the argument for each. Heuristically, we show that the two tasks can be achieved with one single perturbation. 

The UAP is a more challenging scenario, and we can naturally treat IDPs as a special and simple case of UAPs by allowing the UAP to adapt to a specific image. Numerous existing works have attempted to explain IDPs. However, there are limited works that analyze the UAP, which is more challenging to explain due to its ``universal" nature.

\textbf{Deep vs.\ traditional image stenography.}
The primary difference between deep and traditional steganography~\cite{sharda2013image,acharya2013secure} lies in the encoding/decoding mechanism. Traditional image steganography explicitly encodes the secret message with a known predetermined rule, thus how the secret is encoded and decoded is obvious. Deep hiding instead \textit{implicitly} encodes and decodes the message by making the encoder DNN and decoder DNN learn collaboratively for successful hiding and revealing~\cite{baluja2017hiding,baluja2019hiding}. Another difference between the two is that deep steganography has a larger hiding capacity and can hide one (multiple) full-color image(s)~\cite{baluja2017hiding,zhang2020udh}, which makes the DS easily detectable due to the trade-off between secrecy and hiding capacity~\cite{zhu2018hidden,zhang2020udh}. Similarly, detecting the existence of a UAP should not be a challenging task due to its must-have HF property.

\section{Metric quantifying the frequency} 
Fourier transform is one basic tool to perform image frequency analysis. Here, we summarize the main points relevant to this work. Sharp contrast edges in the spatial image are considered as HF content, while smooth or constant patches are LF~\cite{lim1990two}. 
Natural images have the Fourier spectrum concentrated in low-medium frequency range that are in the center of the Fourier image. For performing frequency filtering, we define $X_f = \mathcal{F}^{-1}(f(\mathcal{F}(X), bw))$, where $f$ indicates frequency filtering with the bandwidth $bw$. For high-pass (HP) filtering, $f(z(i,j), bw)$=$z(i,j)$ if $|i-W/2|>=bw/2$ or $|j-H/2|>=bw/2$, otherwise zero; for low-pass (LP) filtering, $f(z(i,j), bw)$=$z(i,j)$ if $|i-W/2|<=bw/2$ and $|j-H/2|<=bw/2$, otherwise zero. $W$ and $H$ are image width and height. Fourier images provide a qualitative presentation for the frequency analysis. No metric has been found to quantify the frequency distribution; to facilitate quantitative cosine similarity analysis in this work, we introduce one simple metric: entropy of the Fourier image $z$, \ie~$E(z)=-\sum_{i}\sum_{j} P(z(i,j)) \text{log}(P(z(i,j)))$ with $ P(z(i,j))$ referring to element probability. Higher entropy indicates more energy being spread to HF regions of $z$, thus indicating the image has more HF content. Note that the entropy is calculated on the transform image $z(i,j)$ instead of the original image.

\section{Methods for USP and UAP}
\subsection{Adopted USP generation method}
\begin{figure}[t]
    \includegraphics[width=\linewidth]{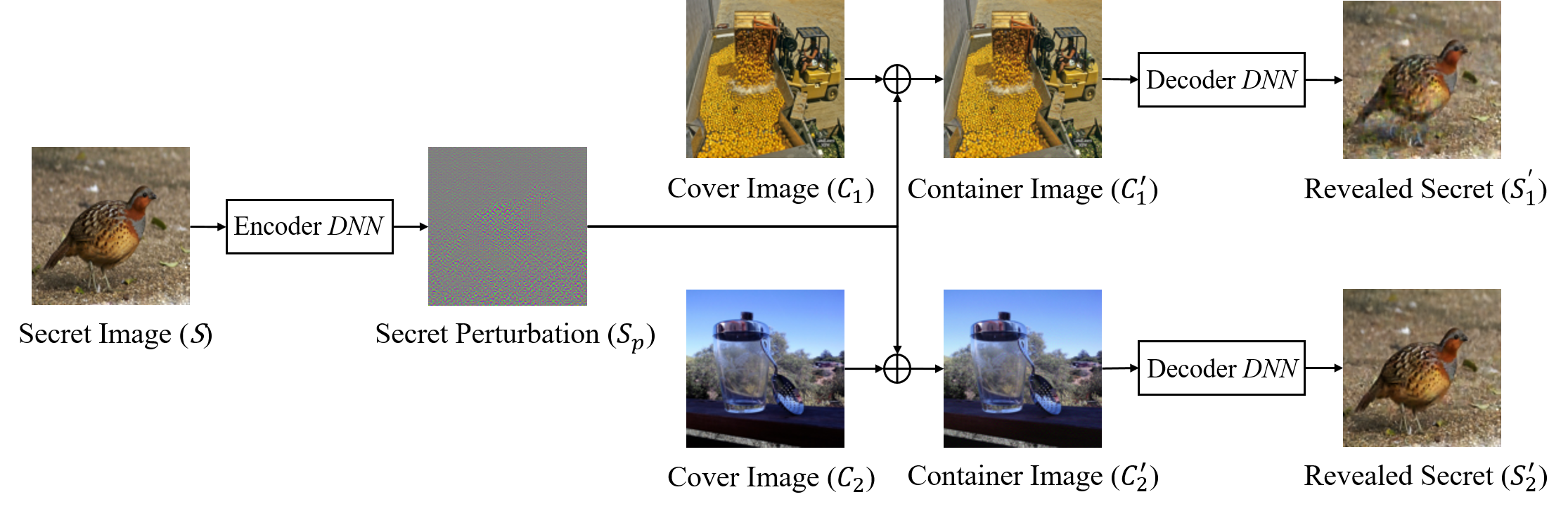}
    \caption{USP generation method. A secret image is encoded to the secret perturbation $S_{p}$, which can be added to random cover images for hiding. We show two different cover images to indicate their random choice.}
    \label{fig:hiding_architecture}
\end{figure}

Our adopted universal secret perturbation (USP) framework~\cite{zhang2020udh} is shown in Figure~\ref{fig:hiding_architecture}. Through a decoder DNN, a secret image $S$ is transformed into a secret perturbation $S_p$, \ie~USP. This $S_p$ can be randomly added to any cover $C$, resulting in container $C'$. From $C'$, the decoder retrieves the hidden secret image $S'$. Following~\cite{zhang2020udh} we use the average pixel discrepancy (APD), defined as the $L_1$-norm of the gap between two images, to measure the hiding and revealing performance.

\begin{figure}[!htbp]
\centering
\small
\scalebox{0.9}{
    \includegraphics[width=0.9\linewidth]{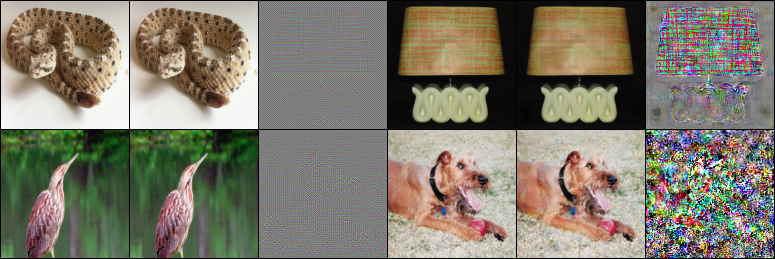}
}
\caption{The first three columns indicate cover image $C$, container image $C'$ and $C'-C$, \ie~$S_{p}$; the next three columns indicate secret image $S$, revealed secret image$S'$ and $S'-S$ respectively. Both $C'-C$ and $S'-S$ are amplified for visualization.}
\label{fig:hiding_main_res}
\end{figure}
Quantitative results evaluated on the ImageNet validation dataset are shown in Table~\ref{tab:hiding_quan}. The two scenarios of IDP and USP are performed with the same procedure in~\cite{zhang2020udh}.
The qualitative results are shown in Figure~\ref{fig:hiding_main_res}, where the difference between $C$ and $C'$ as well as that between $S$ and $S'$ are challenging to identify.
 
\begin{table}[t]
\centering
\small
\caption{Performance comparison for the IDP and USP generation frameworks. We report APD for both cover image (cAPD) and secret image (sAPD). For the secret image, we report the results with the container image (sAPD($C'$)) or only perturbation (sAPD($S_{p}$)) as the input to the decoder network. N/A indicates revealing fails thus not available.}
\scalebox{0.9}{
    \begin{tabular}{ccccccc}
    \toprule
    meta-archs & cAPD  & sAPD ($C'$) & sAPD ($S_{p}$)    \\
    \midrule
    IDP & $2.44$ & $3.42$ & N/A\\
    USP & $2.37$ & $3.52$ & $1.98$ \\
    \bottomrule
    \end{tabular}
    \label{tab:hiding_quan}
}
\end{table}

\subsection{Adopted UAP generation method}
The adopted procedure for generating universal perturbation is illustrated in Algorithm~\ref{alg:simple_uap}, where a differentiable frequency filter $\mathcal{F}$ is adopted to control the frequency of the UAP. We treat the $\mathcal{F}$ as all-frequency pass at this stage, which makes it similar to the UAP algorithm introduced in~\cite{zhang2020understanding,zhang2019cd-uap}. For $\mathcal{L}$, we adopt the widely used negative cross-entropy loss. Except for the image-agnostic nature, this algorithm can be seen adapted from the widely used PGD attack~\cite{madry2017towards,athalye2018obfuscated}. The vanilla UAP~\cite{moosavi2017universal} generation process uses DeepFool~\cite{moosavi2016deepfool} to generate a perturbation to push a single sample over the decision boundary and accumulates those perturbations to the final UAP. The adopted algorithm is different from the vanilla UAP algorithm~\cite{moosavi2017universal} by replacing the relatively cumbersome DeepFool~\cite{moosavi2016deepfool} perturbation optimization with simple batch gradients. ADAM optimizer~\cite{kingma2014adam} is adopted for updating the perturbation values. A similar ADAM based approach has also been adopted for universal adversarial training~\cite{shafahi2020universal}. 

\begin{algorithm}[t]
    \SetAlgoLined
    \DontPrintSemicolon
    \SetKwInput{KwInput}{Input}
    \SetKwInput{KwOutput}{Output}
    \SetKwFunction{FOptim}{Adam}
    \SetKwFunction{Clamp}{Clamp}
    \KwInput{Dataset $\mathcal{X}$, Loss $\mathcal{L}$, Target Model $M$, frequency Filter $\mathcal{F}$, batch size $b$}
    $v \leftarrow 0$ \Comment*[r]{Initialization}
    \For {iteration $=1, \dots, N$}{
        $B \sim \mathcal{X}$: $|B| =  b$ \Comment*[r]{Randomly sample}
        $g_v \leftarrow \underset{x,y \sim B}{\mathds{E}} [\nabla_{v} \mathcal{L}(M(x+\mathcal{F}(v)), y)$] \\
        $v \leftarrow$ \FOptim{$g_v$} \Comment*[r]{Update perturbation} 
        $v \leftarrow$ \Clamp{$v, -\epsilon, \epsilon$} \Comment*[r]{Clamping} 
        }
\caption{Universal attack algorithm}
\label{alg:simple_uap}
\end{algorithm}

Following~\cite{moosavi2017universal,poursaeed2018generative,zhang2020understanding}, we generate the perturbation with $\epsilon=10/255$ on the ImageNet training dataset and evaluate it on the ImageNet validation dataset.
The results for untargeted and targeted UAPs are shown in Table~\ref{tab:uap_quant}. Our simple algorithm achieves high (targeted) fooling ratio.

\begin{table}[t]
\centering
\caption{Performance for untargeted attack (top) with metric fooling ration ($\%$). Performance for the targeted attack (bottom) for target class ``red panda" with metric targeted fooling ratio ($\%$).}
\label{tab:uap_quant}
    \small
    \setlength\tabcolsep{1.2pt}
    \scalebox{0.9}{
    \begin{tabular}{cccccc}
        \toprule
        Method  & AlexNet & GoogleNet  & VGG16  & VGG19  & ResNet152 \\
        \midrule
        Our UAP                        & $94.36$ & $86.03$ & $92.58$ & $94.4$ & $86.67$ \\
        Our HP-UAP   & $91.1$ & $84.4$ & $92.3$ & $90.1$ & $78.4$ \\\midrule
        Our targeted UAP       & $73.77$ & $68.87$ & $81.59$ & $78.67$ & 74.0 \\
        \bottomrule
    \end{tabular}
    }
\end{table}

\section{Explaining the USP induced misalignment}
\begin{figure}[h]
    \centering
    \includegraphics[width=\linewidth]{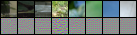}
    \caption{Local patch mapping from corresponding secret image $S$ to secret perturbation $S_p$.}
    \label{fig:hiding_patches}
\end{figure}

In the whole pipeline from $S$ through $S_{p}$ to $S'$, in essence, the role of the $C$ is just like noise. It is counter-intuitive that the pipeline still works well under such large disturbance($||I||\gg||P||$). 
Due to the independent property of $S_p$, we can visualize $S_p$ directly, which is very crucial for qualitatively understanding how the secret image $S$ is encoded in $S_e$~\cite{zhang2020udh}. The visualization in Figure~\ref{fig:hiding_patches} clearly shows that $S_p$ has very HF content.

\textbf{Why does USP have high frequency?}
The decoder network recovers $S'$ from $S_p$ but with the existence of $C$ as a disturbance. Intuitively its role can be decomposed into two parts: distinguishing $S_{p}$ from $C$ in $C'$ and transforming $S_{p}$ to $S'$. 
We conjecture that secret perturbation having high frequency mainly facilitates the role of distinguishing. To verify this, we design a toy task of scale hiding, where we assume/force the encoder to perform a trivial transformation as $S_p = Encoder(S) = S/10$. We then only train the decoder network to perform the inverse up-scaling transformation with the natural $C$ as the disturbance. After the model is trained, we evaluate it in two scenarios: with and without the $C$. The revealing results are present in the supplementary. We observe that the secret image can be recovered reasonably well without the $C$ but fails to work with the $C$. This suggests the transformation $S_{p}$ to $S'$ has been trained well but still is not robust to the disturbance of $C$, which indicates trivial encoding just performing the magnitude change fails. Since natural images $C$ mainly have LF content, it is not surprising that $S_p$ is trained to have HF content, which significantly facilitates the decoder to distinguish $S_p$ from $C$. The decoder network is implicitly trained to ignore LF content in $C$, while transforming the HF $S_p$ back to $S'$. Thus, the revealing performance can be significantly influenced by the image frequency property.

\textbf{Frequency: a key factor for performance.}
We perform analysis with three types of images: artificial flat images with constant values in each RGB channel, natural images, and noise sampled from a uniform distribution of 0 to 1. The results are available in Table~\ref{tab:four_types}. Note that flat images are extremely LF while noise images have HF property. The secret APD performance decreases with the increase of frequency for both secret images and cover images. Since the secret perturbation $S_{p}$ mainly has high frequency, the increase of frequency in the cover images will disrupt more on the $S_{p}$, resulting in the performance to decrease. The task complexity also increases with the increase in the frequency of secret images. Revealing fails when either $S$ or $C$ is random noise.

\section{Explaining the UAP induced misalignment}
We analyze why UAPs tend to have HF property by showing that the target DNN is highly sensitive to HF content.

\textbf{Disentangling Frequency and magnitude.} 
We explore the target DNN's sensitivity to features of different frequencies. Specifically, we analyze the dominance of two independent inputs on the combined output with the cosine similarity $cos$ metric~\cite{zhang2020understanding}. $I$ represents a natural image, while $P$ is an image that extracts the content of a certain frequency range $\omega$ which is one control variable. We normalize $P$ to have uniform magnitude and then multiply it by a new magnitude $m$ which is another control variable. We then calculate $cos(M(I), M(I+P))$ and $cos(M(P), M(I+P))$. For a detailed result, refer to the supplementary, here we summarize the main findings: As expected, a higher magnitude $m$ leads to higher dominance. On the other hand, we find that $\omega$ has an (even more) significant influence on the model prediction. Specifically, higher frequency leads to higher dominance.

\begin{table}[!htbp]
\centering
    \small
    \setlength\tabcolsep{1.2pt}
    \caption{Secret APD performance with three types of images. The rows and columns indicate cover images and secret images, respectively.}
    \scalebox{0.9}{
    \begin{tabular}{ccccc}
        \toprule
        & $S_{Flat}$ & $S_{Natural}$ & $S_{Noise}$ \\
        \midrule
        $C_{Flat}$ & $ 0.34$ & $1.85$ & $48.06$ \\
        $C_{Natural}$ & $1.77$ & $3.52$ & $49.48$ \\
        $C_{Noise}$ & $87.45$ & $98.33$ & $100.47$ \\
        \bottomrule
    \end{tabular}
    \label{tab:four_types}
    }
\end{table}

\begin{figure}[t]
    \centering
    \includegraphics[width=\linewidth]{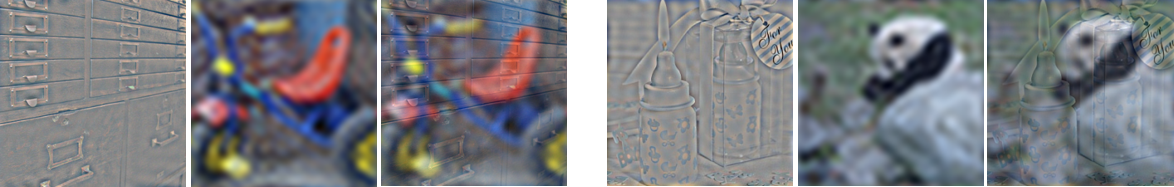}
    \caption{The columns for each image triplet indicate HF image, LF image and hybrid image, respectively.}
    \label{fig:low_high_hybrid}
\end{figure}

\textbf{Hybrid images: HF vs.\ LF.} The target DNN achieves high accuracy and we are interested in finding out whether HF content or LF content dominantly contributes to the success. Note that the targeted DNN has been trained on natural images containing both HF content and LF content and the learning algorithm does not involve any manual intervention to force the model to utilize high or low frequency. Manually forcing the model to specifically learn either LF or HF is possible as performed in~\cite{yin2019fourier}. In contrast to their setup, we evaluate the performance of a normally trained model to filtered images. For a normally trained DNN, we show the usefulness of features with LF or HF content in the natural images as well as explore which side dominates in a hybrid image~\cite{oliva2006hybrid}, which combines the low frequencies of one image with the high frequencies of another. The qualitative results with $bw$ of 20 are available in Figure~\ref{fig:low_high_hybrid}. We observe that a hybrid image visually looks more similar to the LF image. The quantitative results of hybrid images are shown in Table~\ref{tab:comparison_hf_lf_hybrid}. In a hybrid setup, the LF image feature is dominated by the HF one.

\begin{table}[!htbp]
\centering
    \small
    \setlength\tabcolsep{1.2pt}
    
    \caption{Top1 accuracy (\%) for LF, HF, and hybrid images on the ImageNet val dataset evaluated on the VGG19 network. Hybrid HF indicates the accuracy when the HF images labels are chosen as the ground-truth for the Hybrid images. Parallel reasoning applies to Hybrid LF. The columns indicate the bandwidth.}
    \scalebox{0.9}{
    \begin{tabular}{ccccc}
        \toprule
         & 24 & 20 & 16 & 12 \\
        \midrule
        HF   & 23.13 & 31.07 & 41.79 & 53.31 \\
        LF   & 16.07 & 10.62 & 6.14 & 3.04 \\
        Hybrid HF  & 15.95 & 20.39 & 26.54 & 34.31 \\
        Hybrid LF  & 0.87 & 0.52 & 0.32 & 0.21 \\
        \bottomrule
    \end{tabular}
    \label{tab:comparison_hf_lf_hybrid}
    }
\end{table}
The hybrid setup is similar to the universal attack setup because the LF content image is not targeted for any specific HF content image and they are randomly combined. Overall, we observe that the LF image content dominates the human vision, while the HF image content dominates the DNN perception, \ie\ prediction. Given the dominance of the human imperceptible HF content, it is not surprising that the optimization-based UAP with HF property can dominate most natural images for determining the prediction of the target DNN. 

\begin{figure}[!htbp]
    \centering
    \includegraphics[width=\linewidth]{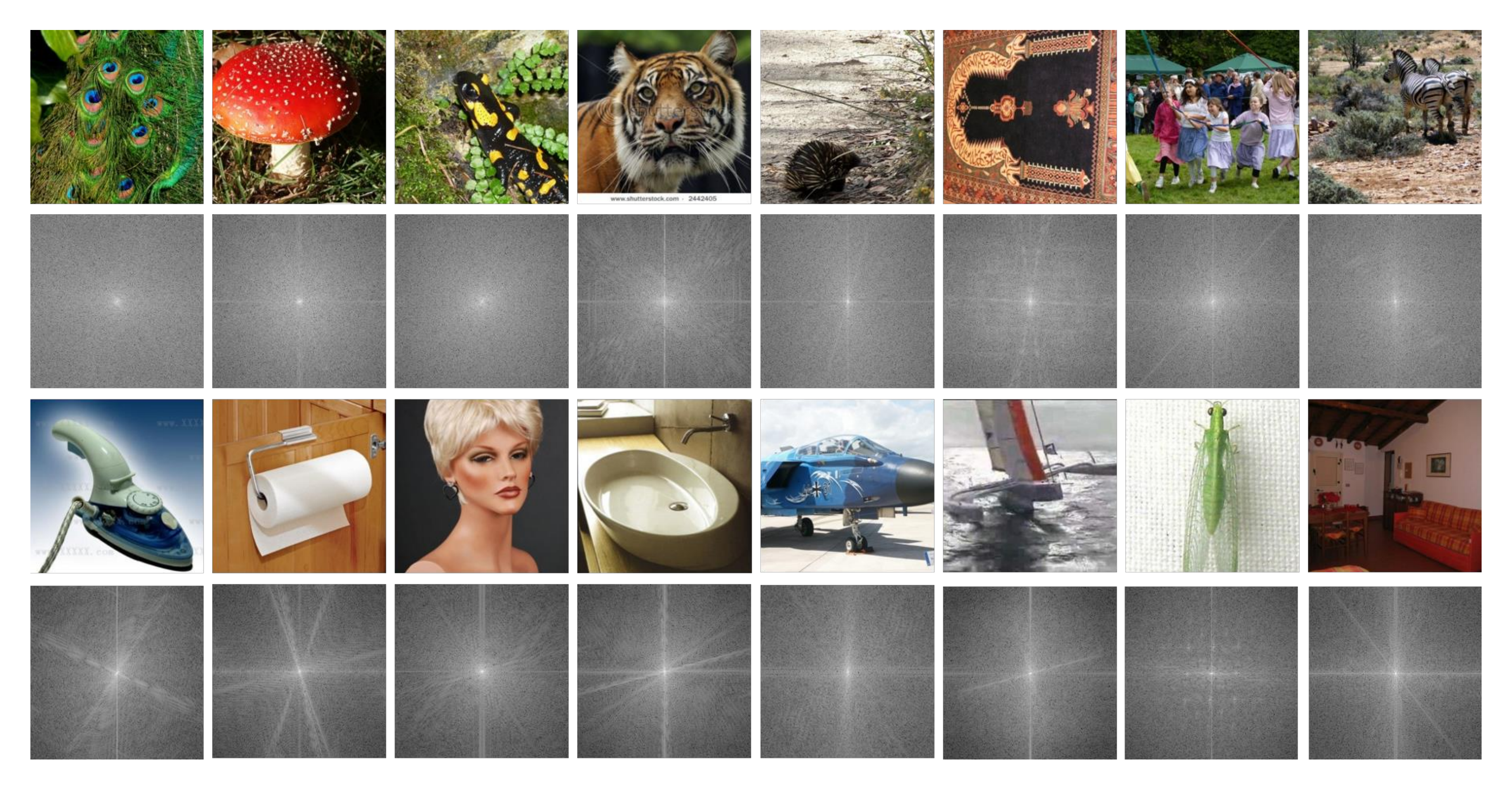}
    \caption{Fourier analysis of representative samples. We randomly choose one sample from 8 top robust classes and non-robust classes to perform Fourier analysis.}
    \label{fig:ft_result}
\end{figure}

\textbf{Frequency: a key factor for class-wise robustness imbalance.}
\label{class_wise}
We randomly choose a targeted class \textit{``red panda"} for performing a universal attack on VGG19. We find that robust classes have a targeted attack success rate of around 40\%, while that for non-robust classes is 100\%. Qualitative results with Fourier analysis are shown in Figure~\ref{fig:ft_result}. 

One interesting observation from the qualitative results is that all the classes with high robustness have repetitive semantic feature patterns, \ie~, HF features, such as the patterns on the feathers of a peacock. The classes with low robustness have LF feature patterns, such as the monotone color of a white washbasin. A Fourier analysis of samples from these classes confirms that robust classes have more HF features, making them more robust to attack. This analysis shows that there are significant class-wise robustness disparity and the key factor that influences its robustness is their frequency. This also provides extra evidence that the DNN is biased towards HF features. Our work is the first to report and analyze this class-wise robustness imbalance.

\section{Joint analysis for two tasks}
\begin{figure}[t]
    \centering
    \parbox{.49\linewidth}{
    \includegraphics[width=\linewidth]{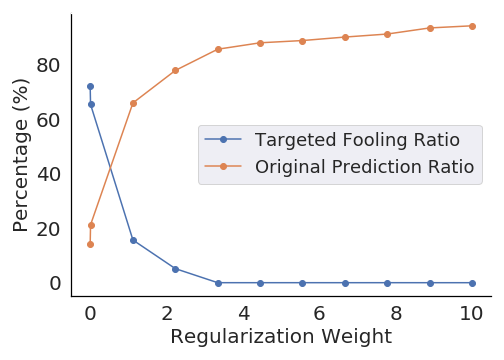}
    \caption{Regularization effect on UAP. Original prediction indicates image samples keeping the same prediction.}
    \label{fig:uap_regularization_quan}
    }\hfill
    \parbox{.459\linewidth}{
    \includegraphics[width=\linewidth]{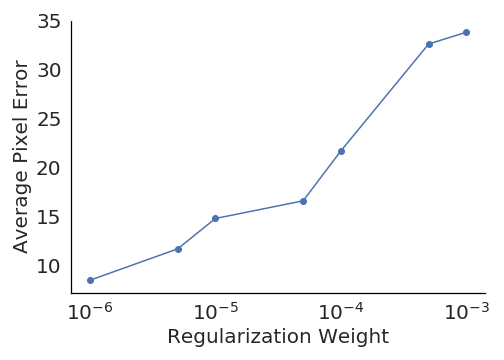}
    \caption{Regularization effect on USP. Secret APD increases with the increase of regularization weight.} 
    \label{fig:hiding_regularization_graph}
    }
\end{figure}

\begin{figure}[t]
    \centering
    \begin{subfigure}[b]{0.49\textwidth}
         \centering
         \includegraphics[width=0.95\linewidth]{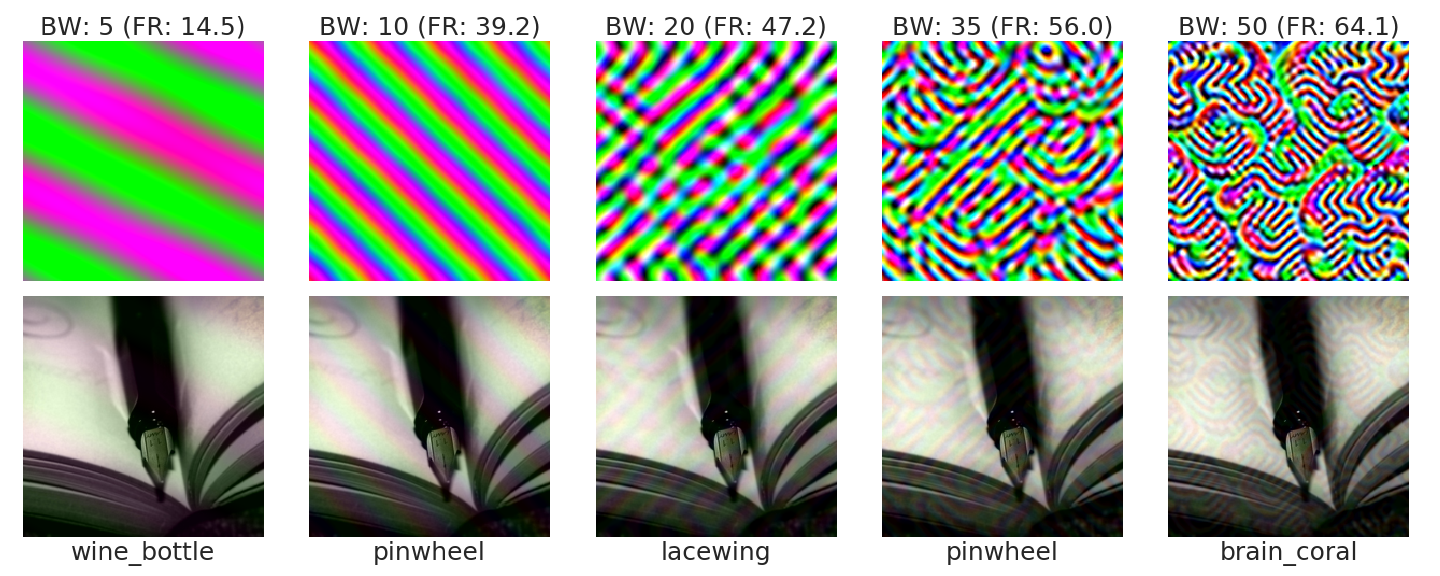} 
    \end{subfigure}
    \begin{subfigure}[b]{0.49\textwidth}
         \centering
         \includegraphics[width=0.95\linewidth]{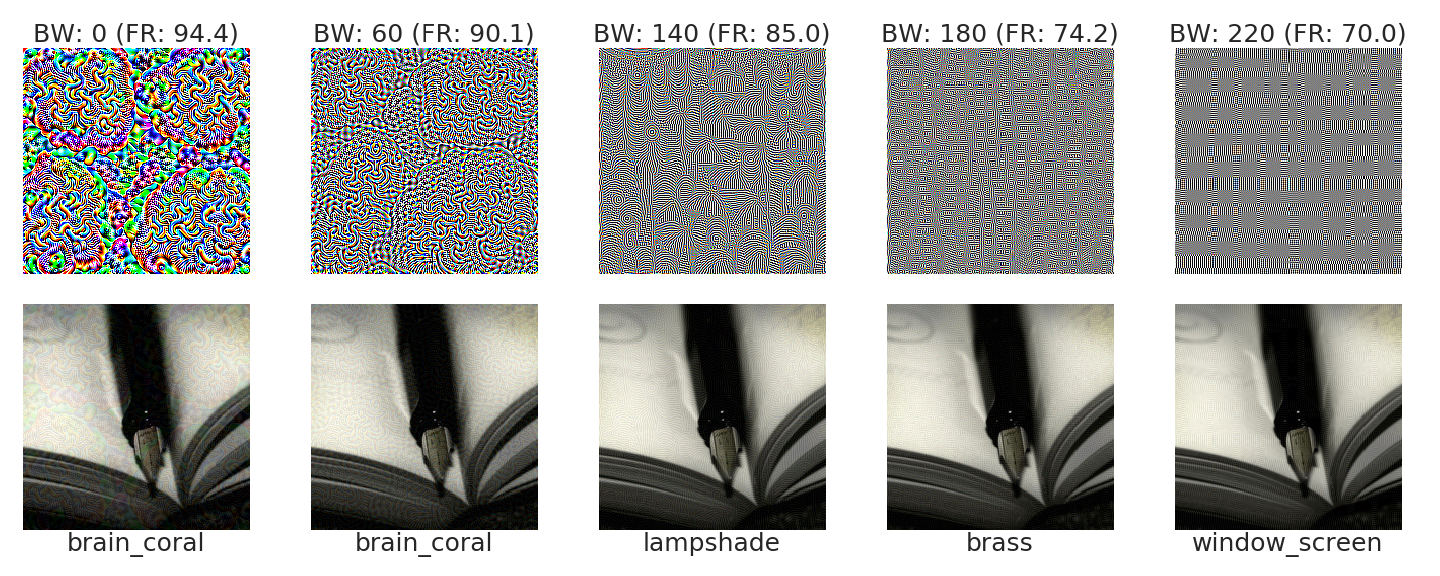} 
    \end{subfigure}
    \caption{Examples for LP UAPs (left) and HP UAPs (right). The first row shows the perturbations for different bandwidths. The used bandwidth (BW) as well as the achieved fooling ratio (FR) are written above the corresponding perturbation. The second row shows the adversarial example with the corresponding predicted class of VGG19 written above. The originally predicted and ground truth class is ``fountain pen".}
    \label{fig:uap_hplp}
\end{figure}

\textbf{Can LF universal perturbation still work?} 
To investigate the behavior of perturbations containing LF features we explore two methods: loss regularization and low-pass filtering. Similar to~\cite{mahendran2015understanding} we add a regularization term to the loss function during universal perturbation generation to force the perturbation to be smooth for both tasks. The results are shown in Figure~\ref{fig:uap_regularization_quan} and Figure~\ref{fig:hiding_regularization_graph}. The results show that regularizing the perturbation to enforce smoothness results in a significant performance drop. 
Higher regularization weight leads to more smooth perturbations (see the supplementary). An LF perturbation can also be enforced by performing an LP filtering to the perturbation before adding the perturbation to the image, for which $\mathcal{F}$ is a differentiable LPF (LP filter) in Algorithm~\ref{alg:simple_uap}. Smoothing the perturbations can remove HF features and lead to lower attack success rates, see Figure~\ref{fig:uap_hplp} (top). Regarding model robustness, we find that \textit{UAP that attacks most images is a strictly high-frequency (HF) phenomenon}. 

\begin{figure*}[t]
    \centering
    \small
    \begin{minipage}{.33\linewidth}
        \centering
        \small
        \includegraphics[width=\linewidth]{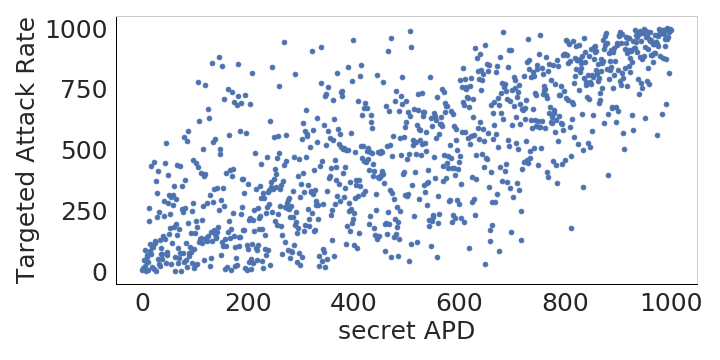}
    \end{minipage}
    \hfill
    \begin{minipage}{.32\linewidth}
        \centering
        \small
        \includegraphics[width=\linewidth]{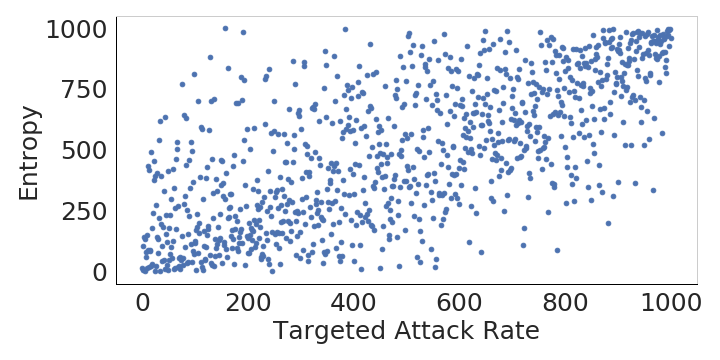}
    \end{minipage}
    \hfill
    \begin{minipage}{.33\linewidth}
        \centering
        \small
        \includegraphics[width=\linewidth]{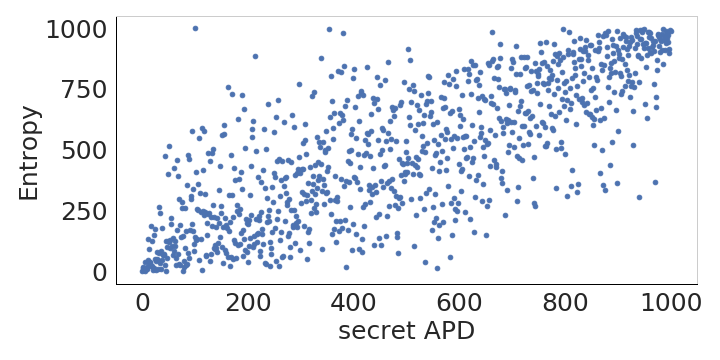}
    \end{minipage}
    \caption{Ranking correlation with three ranking metrics.}
    \label{fig:cross_class_rankings}
\end{figure*}
\textbf{Cross-task cosine similarity analysis for class ranking.} 
We perform a $cos$ analysis between two seemingly unrelated tasks, DS and AA. 
Specifically, the $1000$ ImageNet classes were ranked along the attack success rate metric ($R_1$), secret APD metric ($R_2$) and the Fourier image entropy metric ($R_3$). The ranking plots of $R_1$ over $R_2$, $R_3$ over $R_1$, and $R_3$ over $R_2$ are shown in Figure~\ref{fig:cross_class_rankings}. 
We find that $cos(R_1, R_2)$ is 0.74, indicating high linear correlation for two seemingly unrelated tasks. 
The fact that class robustness is an indicator of the revealing performance in DS task clearly shows that a certain factor exists to link them and we identify this factor to be \textit{frequency}. Note that $R_3$ is the our proposed metric for quantifying the energy distribution (corresponding to each frequency) of Fourier image. $cos(R_1, R_3)$ and $cos(R_2, R_3)$ are 0.68 and 0.77, respectively, attributing the high correlation between $R_1$ ranking and $R_2$ ranking to the \textit{frequency}. 

\section{Feature layer analysis for target DNN} 
\begin{figure*}[t]
    \centering
    \includegraphics[width=0.3\linewidth]{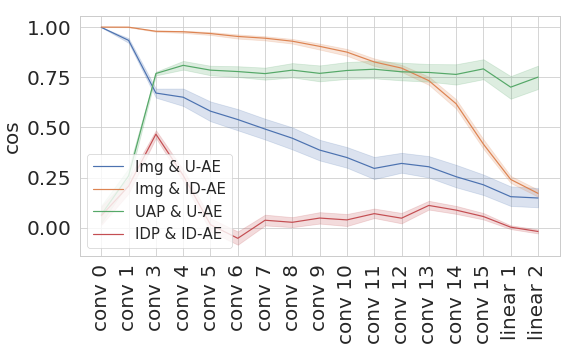}
    \includegraphics[width=0.3\linewidth]{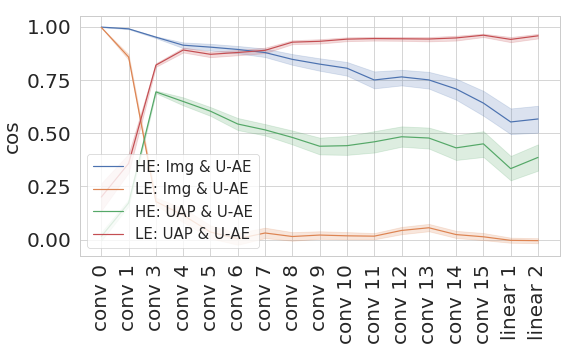}
    \includegraphics[width=0.3\linewidth]{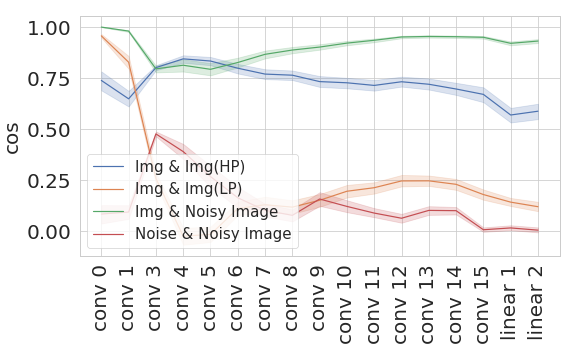}
    \caption{$cos$ analysis on feature layers evaluated on $100$ images. The abbreviations in the legends refer to: image (img), universal/image-dependant adversarial example (U-AE/ID-AE), universal/image-dependant adversarial perturbation (UAP/IDP), high/low entropy (HE/LE), high/low pass (HP/LP) filtered.}
    \label{fig:frequency_analysis}
\end{figure*}
In contrast to prior works with attention only on the DNN output, we analyze feature layers with $cos$ to provide deep insight on generalization and robustness of a target DNN (VGG19). Analysis results are shown in Figure~\ref{fig:frequency_analysis}. 

First, we observe that when $P$ is UAP, $cos(M_{i}(I), M_{i}(I+P))$ is only larger than $cos(M_{i}(P), M(I+P))$ in the first few layers (see Figure~\ref{fig:frequency_analysis} left). In latter layers, $cos(M_{i}(P), M_{i}(I+P))$ is around 0.75, indicating the dominant influence of $P$. Comparing UAP and IDP for $cos(M_{i}(I),  M_{i}(I+P))$, we note that the influence of IDP gets more visible only in the latter layers. $cos(M_{i}(P), M_{i}(I+P))$ for the IDP stays around 0 for all feature layers, indicating the IDP does not have \textit{independent} artificial features as UAP.

Second, with the introduced entropy metric, we explore the influence of the frequency on its robustness to UAP. We find that images of high entropy (HE) (indicating more HF content) are much more robust to UAP on all feature layers, especially on latter layers (see Figure~\ref{fig:frequency_analysis} middle). For example, at layer of $conv6$, $cos(M_{i}(I), M_{i}(I+P))$ is around 0.9 and 0 for images of HE and LE, respectively. The results clearly show that images with more HF content are more robust, which aligns well with the finding that classes with more HF content are more robust. 

Third, comparing $cos(M_{i}(I), M_{i}(HP(I)))$ and $cos(M_{i}(I), M_{i}(LP(I)))$ shows $cos(M_{i}(I), M_{i}(LP(I)))$ is higher only in the first two layers and then significantly lower in latter layers (see Figure~\ref{fig:frequency_analysis} right). It shows that DNN is in general very sensitive to HF instead of LF content, but not for the early layers. When $P$ is noise, $cos(M_{i}(I), M_{i}(I+P))$ first decreases and then increases again, with the conv3 being the most vulnerable to noise. In contrary to adversarial perturbation, the influence of random noise is very limited on latter layers, which provides insight on why DNN is robust to noise.
\begin{figure}[!htbp]
\centering
\small
\scalebox{1}{
    \includegraphics[width=\linewidth]{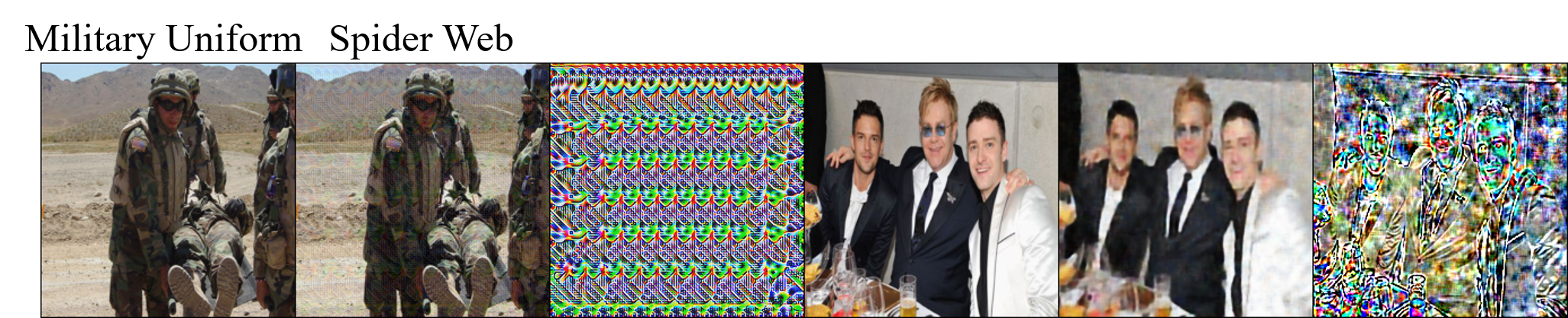}
    }
\caption{Qualitative result of the proposed USAP. The column order is the same as that in the Figure~\ref{fig:hiding_main_res}. The container is misclassified as ``spider web" versus the correct prediction of ``military uniform".}
\label{fig:container_attack}
\end{figure}

\section{Universal secret adversarial perturbation}
We explore whether a single perturbation can fool the DNN for most images while simultaneously containing the secret information. We term it universal secret adversarial perturbation (USAP). Please refer to the supplementary for more details.
Technically, we adopt the same USP generation network, while adding another loss term $NCE(M(C'), y)$ resulting a total loss as $\mathcal{L}(S_{p},S,S',C') = ||S_{p}|| + \beta ||S-S'|| + \gamma NCE(M(C'), y)$ where NCE indicates the negative cross-entropy loss and $y$ indicates the ground-truth label. We set $\beta$ and $\gamma$ to 0.75 and 0.001, respectively. The USAP is constrained to be in the $L_{\infty} = 10/255$. The results are shown in Table~\ref{tab:container_attack} and Figure~\ref{fig:container_attack}, demonstrating a high fooling ratio while containing secret information that can successfully be revealed by the decoder. We are the first to show the existence of such perturbation. 

\begin{table}[!htbp]
\centering
\caption{Performance evaluation of the proposed USAP.}
\label{tab:container_attack}
    \small
    \setlength\tabcolsep{1.2pt}
    \scalebox{0.9}{
    \begin{tabular}{cccccc}
        \toprule
        Metric  & AlexNet & GoogleNet  & VGG16  & VGG19  & ResNet152 \\
        \midrule
        Fooling Ratio                        & $93.8$ & $85.0$ & $92.7$ & $95.8$ & $90.3$ \\
        sAPD   & $13.6$ & $8.9$ & $14.2$ & $11.1$ & $11.9$ \\
        \bottomrule
    \end{tabular}
    }
\end{table}

\section{High-pass UAP}
We create a novel high-pass (HP) universal attack by setting $\mathcal{F}$ to be a differentiable HPF (HP filter) in~Algorithm~\ref{alg:simple_uap}. Overall we observe a performance drop in fooling ratio with increasing $bw$. Results for the HP UAP generated for VGG19 are shown in Figure~\ref{fig:uap_hplp} (bottom). With $bw$ 60, it is much less visible to the human vision and still achieves a fooling ratio of 90.1\%, with only a moderate performance drop compared with the 94.4\% for $bw$ 0 without filtering. 

\section{Conclusion}
This work jointly analyzed AA and DS for the observed misalignment phenomenon and explained their success from the Fourier perspective. With the proposed metric for quantifying frequency distribution, extensive task-specific and cross-task analysis suggests that frequency is a key factor that influences their performance and their success can be attributed to the DNN being highly sensitive to HF content. Our feature layer analysis sheds new light on model generalization and robustness: (a) LF features have more influence on the early layers while HF features have more influence on the later layers; (b) IDP mainly attacks the model on later layers, while UAP attacks most layers with \textit{independent} features. We also proposed two new variants of universal attacks: USAP that simultaneously achieves attack and hiding and HP-UAP that is less visible to the human.

\clearpage
\section{Ethics statement}
Due to security concerns, adversarial attack and deep steganography have become hot topics in recent years. 
We hope that our work will raise awareness of this security concern to the public. 

\bibliography{bib_mixed}

\end{document}